\documentclass[letterpaper]{article} %DO NOT CHANGE THIS
\usepackage{aaai19}  %Required
\usepackage{times}  %Required
\usepackage{helvet}  %Required
\usepackage{courier}  %Required
\usepackage{url}  %Required
\usepackage{graphicx}  %Required
\usepackage{subfigure, multirow, tabularx} % For subfigure
\usepackage{booktabs} % For formal tables
\usepackage{enumitem}
\usepackage{epstopdf}
\usepackage{balance}
\usepackage[noend,ruled,linesnumbered]{algorithm2e} % For algorithm2e
\usepackage{microtype}
\usepackage{url}
\usepackage{wrapfig,lipsum}
\usepackage{amsthm}
\usepackage{amsmath}
\usepackage{amsfonts}
\usepackage{color}

\newtheorem{thm:def}{Definition}
\newtheorem{thm:eg}{Example}
\newtheorem{thm:lem}{Lemma}
\newtheorem{thm:obs}{Observation}
\newtheorem{thm:req}{Requirement}

\newcommand{\nop}[1]{}

\newcommand{\mquote}[1]{{``\emph{#1}''}}

\newcommand{\SynSetMine}{\mbox{\sf SynSetMine}\xspace}

\newcommand{\ie}{\emph{i.e.}\xspace}
\newcommand{\eg}{\emph{e.g.}\xspace}

\title{Mining Entity Synonyms with Efficient Neural Set Generation}

\author{Jiaming Shen$^{\natural}$, Ruiliang Lyu$^{\dag}$, Xiang Ren$^{\ddag}$, Michelle Vanni$^{\diamond}$, Brian Sadler$^{\diamond}$, Jiawei Han$^\natural$ \\
$^{\natural}$Department of Computer Science, University of Illinois Urbana-Champaign, IL, USA \\
$^{\dag}$Department of Electronic Engineering, Shanghai Jiao Tong University, China $\quad$ $^{\diamond}$U.S. Army Research Laboratory, MD, USA\\
$^{\ddag}$Department of Computer Science, University of Southern California, CA, USA \\
$^{\natural}$\{js2, hanj\}@illinois.edu $\quad$ $^{\dag}$lvruiliang-ele@sjtu.edu.cn $\quad$ $^{\ddag}$xiangren@usc.edu \\
$^{\diamond}$\{michelle.t.vanni.civ, brian.m.sadler6.civ\}@mail.mil
}

\begin{document}
\maketitle

\begin{abstract}
	%!TEX root = main.tex
% UTF-8 encoding

Mining entity synonym sets (\ie, sets of terms referring to the same entity) is an important task for many entity-leveraging applications. 
Previous work either rank terms based on their similarity to a given query term, or treats the problem as a two-phase task (\ie, detecting synonymy pairs, followed by organizing these pairs into synonym sets). 
However, these approaches fail to model the \textit{holistic} semantics of a set and suffer from the error propagation issue.
Here we propose a new framework, named \SynSetMine, that efficiently generates entity synonym sets from a given vocabulary, using example sets from external knowledge bases as distant supervision.
\SynSetMine consists of two novel modules:
(1) a set-instance classifier that jointly learns how to represent a permutation invariant synonym set and whether to include a new instance (\ie, a term) into the set, and 
(2) a set generation algorithm that enumerates the vocabulary only once and applies the learned set-instance classifier to detect all entity synonym sets in it. 
Experiments on three real datasets from different domains demonstrate both effectiveness and efficiency of \SynSetMine for mining entity synonym sets. 
\end{abstract}

%!TEX root = main.tex
% UTF-8 encoding
\section{Introduction}\label{sec:intro}

\begin{figure*}[!t]
  \centering
  \centerline{\includegraphics[width=1.0\textwidth]{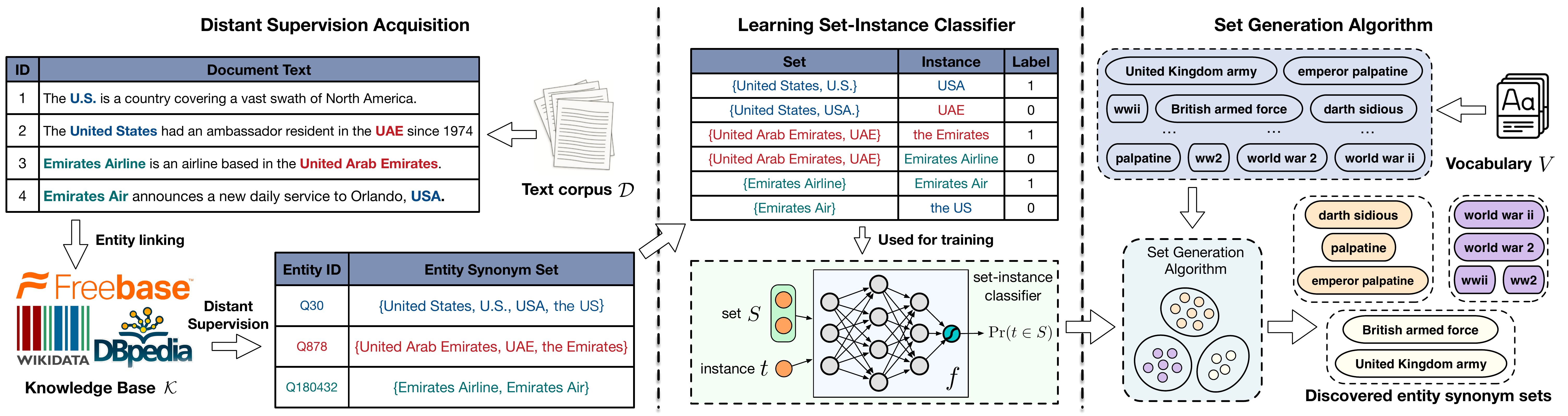}}
  \caption{\SynSetMine Framework Overview.}
  \label{fig:framework}
\end{figure*}

An \emph{entity synonym set} is a set of terms (\ie, words or phrases) that refer to the same real-world entity.
For instance, \{\mquote{USA}, \mquote{United States}, \mquote{U.S.}\} is an entity synonym set as all terms in it refer to the same country.
Entity synonym set discovery can benefit a wide range of applications such as web search \cite{Cheng2012EntitySF}, question answering \cite{Zhou2013ImprovingQR}, and taxonomy construction \cite{anh2015incorporating}.
Take the query \mquote{Emirates Airline U.S. to UAE} as an example, understanding \mquote{U.S.} refers to \mquote{United States} and \mquote{UAE} stands for \mquote{United Arab Emirates} is crucial for \nop{any system to return good search results}an intelligent system to satisfy the user information need.

One line of work for entity synonym sets discovery takes a \emph{ranking plus pruning} approach.
Given a query term referring to one entity, it first ranks all candidate terms based on their probabilities of referring to the same entity and then prunes the rank list into an output set.
By treating each term in a vocabulary as a query, this approach can finally output all the entity synonym sets in the vocabulary.
A variety of features are extracted, including corpus-level statistics \cite{Turney2001MiningTW}, textual patterns \cite{Nakashole2012PATTYAT}, or query contexts \cite{Chakrabarti2012AFF}, from different data sources (\eg, query log \cite{Chaudhuri2009ExploitingWS}, web table \cite{He2016AutomaticDO}, and raw text corpus \cite{Qu2017AutomaticSD}) to calculate the above probabilities.
However, this approach treats each candidate term separately and computes its probability of referring to the query entity independently.
As a result, it ignores relations among candidate terms which could have helped to improve the quality of discovered synonym sets.
Furthermore, as shown in \cite{Ren2015SynonymDF}, the number of true synonyms varies a lot across different entities and thus converting a rank list into a set itself is a non-trivial problem and can be error-prone.

Another line of work divides the synonym set discovery problem into two sequential subtasks: (1) \emph{synonymy detection} (\ie, finding term pairs of synonymy relation), and (2) \emph{synonymy organization} (\ie, aggregating synonymous term pairs into synonym sets).
Methods developed for synonymy detection leverage textual patterns \cite{Wang2012ExploringPI} and distributional word representations \cite{Shwartz2016CogALexVST} to train a classifier that predicts whether two terms hold the synonymy relation.
Then, those predicted term pairs form a synonymy graph on which different graph clustering algorithms are applied \cite{Hope2013MaxMaxAG,Oliveira2014ECOAO,Ustalov2017WatsetAI}.
This approach is able to capture relations among candidate terms and returns all entity synonym sets in the vocabulary.
However, these two-phase methods only use training data in their first phase and cannot leverage training signals in the second phase.
Furthermore, the detected synonymous term pairs are usually fixed during synonymy organization and there is no feedback from the second phase to the first, which causes the error accumulation problem.

In this work, we propose a new framework, \SynSetMine, which leverages existing synonym sets from a knowledge base as \emph{distant supervision} and extracts more synonym sets not in knowledge bases from massive raw text corpus.
Specifically, \SynSetMine first applies an entity linker to map in-corpus text (\ie entity mentions) to entities in the knowledge base.
Then, it groups all mentions mapped to the same entity (with the same unique id) into an entity synonym set, which provides supervision signals.
As these ``training" synonym sets are created automatically and without any human effort, we refer to them as distant supervision.

To effectively leverage distant supervision signals, \SynSetMine consists of two novel modules.
First, we train a set-instance classifier which jointly learns how to represent an entity synonym set and whether to include a new instance (\ie, a term) into the set.
This set-instance classifier can model a set holistically, instead of decomposing it into separated pairs.
As a result, it effectively captures relations among candidate terms and directly leverages supervision signals from the set structure.
Second, we design an efficient set generation algorithm that applies the learned set-instance classifier to discover new entity synonym sets.
Given a vocabulary, this algorithm processes each term in it one at a time and maintains a pool of all detected sets.
For each term in the vocabulary, the algorithm applies the set-instance classifier to determine whether and which previous detected set this term should reside in.
If no matching set can be found, a new set is formed, consisting of this single term, and added into the pool of detected sets.
As it only enumerates the vocabulary once to generate all synonym sets, this algorithm is efficient.

\smallskip 
\noindent
\textbf{Contributions.} This study makes three contributions:
(1) a novel framework, \SynSetMine, is proposed that leverages distant supervision for entity synonym set mining;
(2) a set-instance classifier is designed to model entity synonym sets holistically and is integrated into a set generation algorithm to discover new synonym sets efficiently; and
(3) extensive experiments conducted on three real-world datasets from different domains show the effectiveness of the method.

%!TEX root = main.tex
% UTF-8 encoding
\section{Problem Formulation}\label{sec:problem}
We first elaborate on some important concepts and then formulate the problem.

\smallskip
\noindent \textbf{Entity Synonym Set.} An entity synonym set is a set of terms (\ie, words or phrases) that refer to the same real-world entity.

\smallskip
\noindent \textbf{Knowledge Base.} A knowledge base consists of many facts about a set of entities. In this work, we focus on one particular type of facts: entity synonym. 
For some entities, their synonyms are manually curated and stored in a knowledge base. 
The knowledge base provides such training signals to help discover more entity synonym sets.

\smallskip
\noindent \textbf{Problem Definition.} 
Given a text corpus $\mathcal{D}$, a vocabulary $V$ (\ie, a list of terms) derived from $\mathcal{D}$, and a knowledge base $\mathcal{K}$, the task of \emph{mining entity synonym set} aims to discover all entity synonym sets consisting of terms in $V$, based on the information extracted from $\mathcal{D}$ and $\mathcal{K}$. 
%!TEX root = main.tex
% UTF-8 encoding
\section{The \SynSetMine Framework}\label{sec:method}

Our proposed \SynSetMine framework consists of three main steps (Figure \ref{fig:framework}):
(1) an entity linker is used to map in-corpus text (\ie, entity mentions) to entities in the given knowledge base, which provides some training entity synonym sets as distant supervision;
(2) a classifier is constructed to determine whether a term should be included into the synonym set, based on the above distant supervision; and
(3) the learned classifier is integrated into a set generation algorithm which outputs all term clusters in the vocabulary as detected entity synonym sets.

%% Step 1: Distant Supervision Acquisition
\subsection{Distant Supervision Acquisition}
A knowledge base contains a collection of curated entities with their known synonyms.
These entities can provide distant supervision signals to help discover more entity synonym sets that are not in the knowledge base from raw text corpus.
To automatically acquire the distant supervision, we first apply an existing entity linker such as DBpedia Spotlight \cite{Mendes2011DBpediaSS} which directly maps in-corpus text (\ie, entity mentions) to entities in the knowledge base.
However, most entity linkers are not perfect and heavily rely on the string-level features which could introduce additional noise, as shown in \cite{Shen2018EntitySS} and the example below. 
Therefore, we follow the same procedure in \cite{Qu2017AutomaticSD} to reduce the linking errors.
Specifically, for each entity mention and its linked entity, if the mention surface string is not in that entity's synonym set (in the knowledge base), we remove the link between them.
Finally, we group all entity mentions that linked to the same entity as a training entity synonym set, and collect all synonym sets from the linked corpus as distant supervision.
\begin{thm:eg}
Given a sentence ``The U.S. Steel, located in Pennsylvania, USA, is a leading steel producer in America'', an entity linker may first map ``U.S. Steel'', ``USA'' and ``America'' to the entity ``\textsc{United States}''.
Then, we find that the synonym set of entity ``\textsc{United States}'', retrieved from knowledge base, does not contain the surface string ``U.S. Steel''.
Therefore, we remove the link between them and group the remaining two entity mentions into an entity synonym set \{``USA'', ``America''\}.
\end{thm:eg}

%% Step 2: Learning Set-Instance Classifier
\subsection{Learning Set-Instance Classifier}
After obtaining distant supervision, we train a set-instance classifier, denoted as $f(S,t)$, to determine whether a synonym set $S$ should include an instance $t$ (\ie, a term).

\subsubsection{Set-Instance Classifier Architecture.}
\begin{figure}[!t]
  \centering
  \centerline{\includegraphics[width=0.48\textwidth]{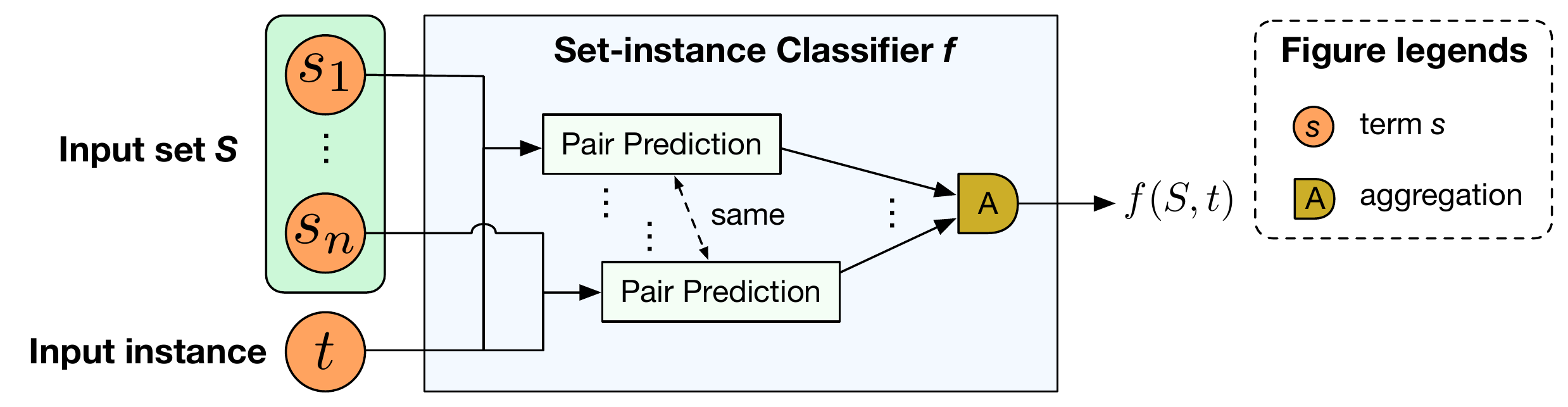}}
  \caption{Aggregating instance-instance pair prediction results into set-instance prediction result.}
  \label{fig:pair-pair}
\end{figure}

One key requirement of the set-instance classifier $f(S,t)$ is that its output should be invariant to the ordering of elements in set $S$.
An intuitive way to achieve such \emph{permutation invariance} property is to decompose the set-instance prediction into multiple instance-instance pair predictions, as shown in Figure \ref{fig:pair-pair}.
Each pair prediction decides whether two instances are synonyms, and all pair prediction results are finally aggregated into the set-instance prediction result.
However, this method completely ignores the relations between elements in set $S$.

\begin{figure}[!t]
  \centering
  \centerline{\includegraphics[width=0.48\textwidth]{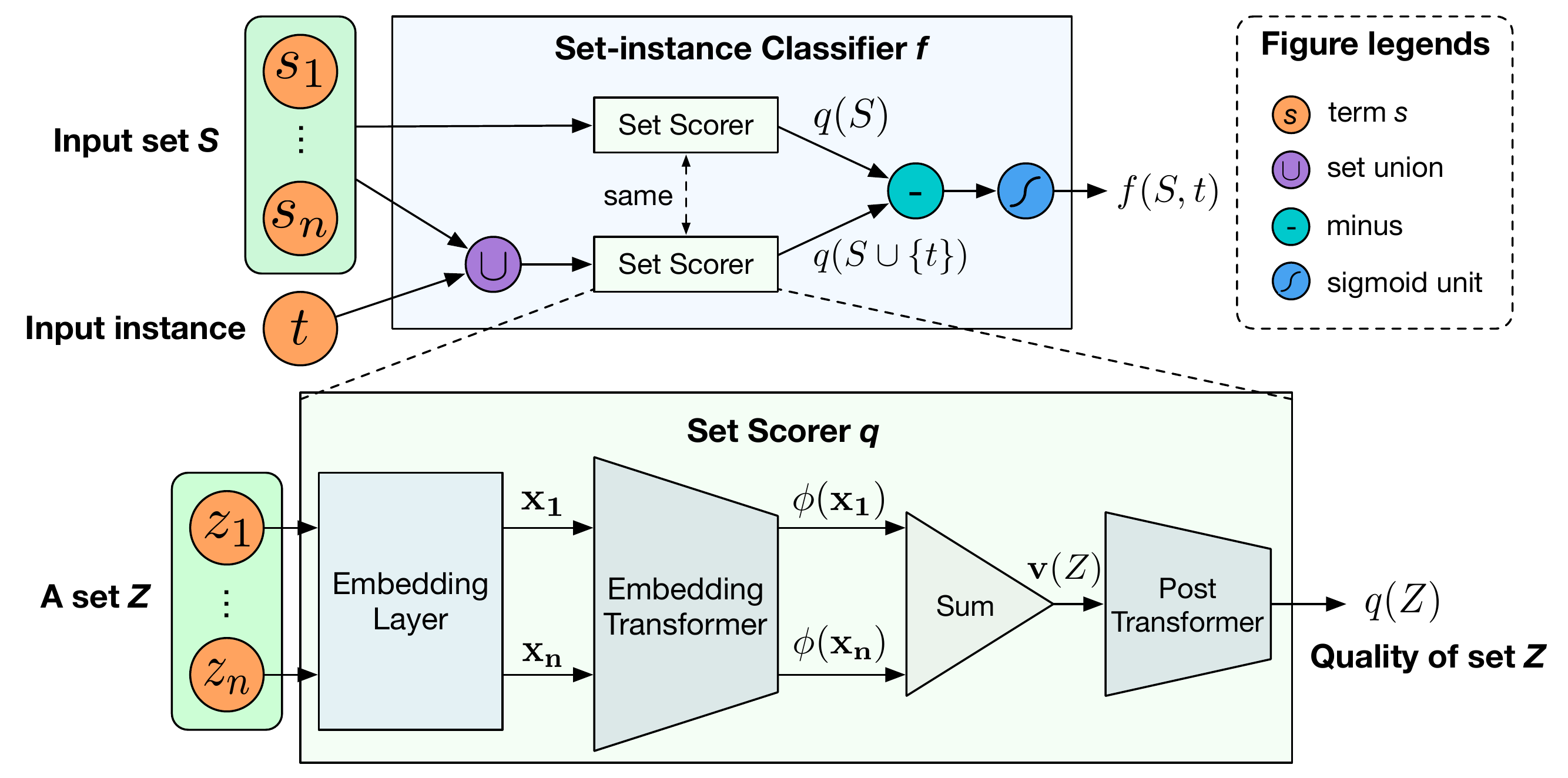}}
  \caption{Architecture of our set-instance classifier.}
  \label{fig:architecture}
\end{figure}

In this work, we are inspired by \cite{Zaheer2017DeepS} and design a neural network architecture that directly learns to represent the permutation invariant set.
The architecture of our set-instance classifier is shown in Figure \ref{fig:architecture}.
The bottom part of Figure \ref{fig:architecture} shows a set scorer $q(\cdot)$ which takes a set $Z$ as input, and returns a quality score $q(Z)$ that measures how complete and coherent this set $Z$ is.
Given a synonym set $S$ and an instance term $t$, our set-instance classifier $f(\cdot)$ first applies the set scorer to obtain the quality score of input set $S$ (\ie, $q(S)$).
Then, we add the instance $t$ into the set and apply the set scorer again to obtain the quality score of $S \cup \{t\}$.
Finally, we calculate the difference between these two quality scores, and transform this score difference into the probability using a sigmoid unit as follows:
\begin{equation}
\small
\Pr(t \in S) = f(S, t) = \phi \left(q(S \cup \{t\}) - q(S) \right),
\end{equation}
where $\phi(x) = \frac{1}{1+e^{-x}}$ is the sigmoid function.

Given a collection of $m$ set-instance pairs $\{(S_{i}, t_{i}) |_{i=1}^{m}\}$ with their corresponding labels $\{y_{i}|_{i=1}^{m}\}$, we learn the set-instance classifier using the log-loss as follows:
\begin{equation}
\small
\mathcal{L}(f) = \sum_{i=1}^{m} -y_{i} \log (f(S_{i}, t_{i})) - (1-y_{i}) \log(1-f(S_{i}, t_{i})),
\end{equation}
where $y_{i}$ equals to 1 if $t_{i} \in S_{i}$ and equals to 0 otherwise.

Note that if the set scorer $q(\cdot)$ is invariant to the ordering of elements in its input set, our set-instance classifier built on it will naturally satisfy the permutation invariance property.
Following we first describe the set scorer architecture (the bottom part of Figure \ref{fig:architecture}) and then discuss how to generate set-instance pairs from distant supervision.

%% Figure: Aggregation of pair prediction architecture

%% Figure: Set-Instance Classifier Architecture

\subsubsection{Set Scorer Architecture.}
Given a set of terms $Z = \{z_1, \dots, z_n\}$, the set scorer first passes each term $z_{i}$ into an embedding layer and obtains its embedding $\mathbf{x_{i}}$.
Then, an ``embedding transformer'' $\phi(\cdot)$ is applied to transform the raw embedding to a new term representation $\phi(\mathbf{x_{i}})$.
After that, we sum all transformed term representations and obtain the ``raw'' set representation $\mathbf{v}(Z) = \sum_{i=1}^{n}\phi(\mathbf{x_{i}})$.
Finally, we feed this set representation into the ``post-transformer'' which outputs the final set quality score.
Since the summation operation is commutative, the ``raw'' set representation $\mathbf{v}(Z)$ is invariant to the ordering of elements in $Z$ and so is the set-scorer $q$.

In this work, we initialize the embedding layer using term embeddings pre-trained on the given corpus.
We instantiate the ``embedding transformer'' using a fully connected neural network with two hidden layers.
For the ``post transformer'', we construct it using another fully connected neural network with three hidden layers.
We demonstrate the necessity of these two transformers and study how the size of hidden layers may influence model performance in later experiments.

\subsubsection{Generation of Training Set-Instance Pairs.}\label{subsubsec:generation-set-instance-pair}
To train the set-instance classifier $f(S,t)$, we need to first generate a collection of training set-instance pairs from training synonym sets.
For each entity synonym set $ES$, we randomly holdout one instance $t^{pos} \in ES$ and construct a positive set-instance sample $(S^{pos}, t^{pos})$ where $S^{pos} = ES \setminus \{t^{pos}\}$.
Then, for each positive sample $(S^{pos}, t^{pos})$, we generate $K$ negative samples by selecting $K$ negative instances $t^{neg}_{i}|_{i=1}^{K}$ and pair each of them with $S^{pos}$.
To generate each negative instance $t^{neg}_{i}$, we can either randomly choose a term from the vocabulary $V$ (denoted as \texttt{complete-random} strategy); select a term that shares same token with some string in $S^{pos}$ (denoted as \texttt{share-token} strategy), or combine these two methods (denoted as \texttt{mixture} strategy).
We study how this negative sample size $K$ and the sampling strategy influence the model performance in later experiments.

%% Step 3: Set Generation Algorithm
\subsection{Set Generation Algorithm}
We present our designed set generation algorithm for mining entity synonym set in Algorithm~\ref{alg:algorithm1}.
This algorithm takes the above learned set-instance classifier, a vocabulary $V$, and a probability threshold $\theta$ as input, and clusters all terms in the vocabulary into entity synonym sets.
Specifically, this algorithm enumerates the vocabulary $V$ once and maintains a pool of all detected sets $\mathcal{C}$.
For each term $s_i \in V$, it applies the set-instance classifier $f$ to calculate the probability of adding this term into each detected set in $\mathcal{C}$, and finds the best set $C_{j}$ that has the largest probability,
If this probability value passes the threshold $\theta$, we will add $s_i$ into set $C_{j}$.
Otherwise, we create a new set $\{s_i\}$ with this single term and add it into the set pool $\mathcal{C}$.
The entire algorithm stops after one pass of the vocabulary and returns all detected sets in $\mathcal{C}$.
Note that we do not need to specify the number of entity synonym sets in the vocabulary and our set generation algorithm will determine this value on its own.
In this work, we simply set the probability threshold $\theta$ to be 0.5 and study its influence on clustering performance below.

\begin{algorithm}[!t]
\small
  \caption{Set Generation Algorithm}
  \label{alg:algorithm1}
  \KwIn{
    A set-instance classifier $f$; An input vocabulary $V = (s_1, s_2, \dots, s_{|V|})$; A threshold $\theta \in [0,1]$.
  }
  \KwOut{$m$ entity synonym sets $\mathcal{C} = [C_1, C_2, \dots, C_{m}]$ where $C_{i} \subseteq V$, $\cup_{i=1}^{m} C_{i} = V$, $C_{i} \cap C_{j} = \emptyset, \forall i \neq j$.
  }
  $\mathcal{C} \gets [\{s_1\}]$; // initialize the first single-element cluster\;
  \For{$i$ from 2 to $|V|$} {
  	$best\underline{\hspace{0.05in}}score$ = 0\;
  	$best\underline{\hspace{0.05in}}j$ = 1\;
	\For{$j$ from 1 to $|\mathcal{C}|$}{
		\If{$ f(C_{j}, s_{i}) > best\underline{\hspace{0.05in}}score$} {
			$best\underline{\hspace{0.05in}}score \gets f(C_{j}, s_{i})$\;
			$best\underline{\hspace{0.05in}}j \gets j$\;
		}
	}
	\If{$best\underline{\hspace{0.05in}}score > \theta$} {
		$C_{best\underline{\hspace{0.05in}}j}.add(s_{i})$\;
	}
	\Else{
		$\mathcal{C}$.append(\{$s_i$\}); //add a new cluster into the output\;
	}
  }
  Return $\mathcal{C}$\;
\end{algorithm}

\subsubsection{Complexity Analysis.}
To compare with the clustering algorithm that requires the input cluster number, we suppose our algorithm will eventually return $K$ clusters.
Then, our algorithm applies the set-instance classifier at most $O(|V| \cdot K)$ times, where $|V|$ denotes the vocabulary size.
Most computation efforts of set-instance classifier are the matrix multiplication in two transformers, which can be accelerated by GPU.
As a comparison, the time complexity of $k$-means algorithm is $O(|V| \cdot K \cdot I)$ where $I$ denotes the number of iterations, 
and for most supervised clustering methods (such as $SVM^{cluster}$ \cite{Finley2005SupervisedCW}) and two-phase methods (such as WATSET \cite{Ustalov2017WatsetAI}), their time complexity is $O(|V|^{2})$ as they need to explicitly calculate all pairwise similarities. 
Finally, we emphasize that we can further parallelize lines 5-8 in Algorithm~\ref{alg:algorithm1} by grouping all set-instance pairs $\{ (C_j, s_i) | j = 1, \cdots, |\mathcal{C}|\}$ into a batch and applying the set-instance classifier only once.
In practice, this strategy can significantly reduce the running time.

%!TEX root = main.tex
% UTF-8 encoding
\section{Experiments}\label{sec:exp}
In this section, we first describe our experimental setup, and then report the experimental results. 
Finally, we analyze each component of \SynSetMine in more details and show several concrete case studies. 
Our model implementation is available at: \url{https://github.com/mickeystroller/SynSetMine-pytorch}.

%% Experimental Setup
\subsection{Experimental Setup}

\smallskip
\noindent 
\textbf{Datasets.}$~~$
We evaluate \SynSetMine on three public benchmark datasets used in \cite{Qu2017AutomaticSD}:
\begin{enumerate}
\item \textbf{Wiki} contains 100K articles in the Wikipedia. We use Freebase\footnote{\scriptsize \url{https://developers.google.com/freebase/}} as the knowledge base. 
\item \textbf{NYT} includes about 119K news articles from 2013 New York Times. We use Freebase as the knowledge base.
\item \textbf{PubMed} contains around 1.5M paper abstracts from PubMed\footnote{\scriptsize \url{https://www.ncbi.nlm.nih.gov/pubmed}}. We select UMLS\footnote{\scriptsize \url{https://www.nlm.nih.gov/research/umls/}} as the knowledge base. 
\end{enumerate}
For Wiki and NYT datasets, DBpedia Spotlight\footnote{\scriptsize \url{https://github.com/dbpedia-spotlight/dbpedia-spotlight}} is used as the entity linker, and for PubMed, we apply PubTator\footnote{\scriptsize \url{https://www.ncbi.nlm.nih.gov/CBBresearch/Lu/Demo/PubTator/}} as the entity linker. 
After the linking step, we randomly sample a portion of linked entities as test entities and treat the remaining as training entities. 
Note that there is no overlapping between training vocabulary and testing vocabulary, which makes the evaluation more realistic but also more challenging. 
The statistics of these datasets are listed in Table \ref{tab:datasets}. 
All datasets are available at: \url{http://bit.ly/SynSetMine-dataset}.
 
%% Table: dataset
\begin{table}[!t]
    \centering
    \caption{Datasets Statistics.}\label{tab:datasets}
\scalebox{0.8}{
    \begin{tabular}{cccc}
        \toprule
        \textbf{Dataset}    & \textbf{Wiki} &  \textbf{NYT} & \textbf{PubMed} \\
        \midrule
        \#Documents       &100,000      &118,664      &1,554,433                   \\
	\#Sentences       &6,839,331      &3,002,123      &15,051,203                    \\
	\midrule
	\#Terms in \emph{train} &8,731       &2,600      &72,627                  \\
	\#Synonym sets in \emph{train} &4,359       &1,273      &28,600                  \\
	\midrule
	\#Terms in \emph{test} &891       &389      &1,743                  \\
	\#Synonym sets in \emph{test} &256       &117      &250                  \\	
	\bottomrule
    \end{tabular}
}
\end{table}

\begin{table*}[!t]
\centering
\caption{Quantitative results of entity synonym set mining. All metrics are in percentage scale. We run all methods except L2C five times and report the averaged score with standard deviation. Due to the bad scalability of L2C, we have not obtain its results on PubMed dataset within 120h, and indicate this using ``--'' mark.}
\scalebox{0.70}{
        \begin{tabular}{c|ccc|ccc|ccc}
            \toprule
            \multirow{2}{*}{\textbf{Method}} & \multicolumn{3}{c}{\textbf{Wiki}} & \multicolumn{3}{c}{\textbf{NYT}} & \multicolumn{3}{c}{\textbf{PubMed}}  \\
            \cmidrule{2-10}
                           & ARI ($\pm$std) & FMI ($\pm$std) & NMI ($\pm$std)        & ARI ($\pm$std) & FMI ($\pm$std) & NMI ($\pm$std)      & ARI ($\pm$std) & FMI ($\pm$std) & NMI ($\pm$std) \\
                          \midrule
                         Kmeans & 34.35 ($\pm$1.06) & 35.47 ($\pm$0.96) & 86.98 ($\pm$0.27) & 28.87 ($\pm$1.98) & 30.85 ($\pm$1.76) & 83.71 ($\pm$0.57) & 48.68 ($\pm$1.93) & 49.86 ($\pm$1.79) & 88.08 ($\pm$0.45) \\
                         Louvain & 42.25 ($\pm$0.00) & 46.48 ($\pm$0.00) & 92.58 ($\pm$0.00) & 21.83 ($\pm$0.00) & 30.58 ($\pm$0.00) & 90.13 ($\pm$0.00) & 46.58 ($\pm$0.00) & 52.76 ($\pm$0.00) & 90.46 ($\pm$0.00) \\
                         SetExpan+Louvain & 44.78 ($\pm$0.28) & 44.95 ($\pm$0.28) & 92.12 ($\pm$0.02) & 43.92 ($\pm$0.90) & 44.31 ($\pm$0.93) & 90.34 ($\pm$0.11) & 58.91 ($\pm$0.08) & 61.87($\pm$0.07) & 92.23 ($\pm$0.15) \\
			COP-Kmeans & 38.80 ($\pm$0.51) & 39.96 ($\pm$0.49) & 90.31 ($\pm$0.15) & 33.80 ($\pm$1.94) & 34.57 ($\pm$2.06) & 87.92 ($\pm$0.30) & 49.12 ($\pm$0.85) & 51.92 ($\pm$0.83) & 89.91 ($\pm$0.15) \\
			\midrule
			SVM+Louvain & 6.03 ($\pm$0.73) & 7.75 ($\pm$0.81) & 25.43 ($\pm$0.13) & 3.64 ($\pm$0.42) & 5.10 ($\pm$0.39) & 21.02 ($\pm$0.27) & 7.76 ($\pm$0.96) & 8.79 ($\pm$1.03) & 31.08 ($\pm$0.34)\\			
			L2C & 12.87 ($\pm$0.22) & 19.90 ($\pm$0.24) & 73.47 ($\pm$0.29) & 12.71 ($\pm$0.89) & 16.66 ($\pm$0.68) & 70.23 ($\pm$1.20) & -- & -- & -- \\			
			\midrule
			\SynSetMine & \textbf{56.43 ($\pm$1.31)} & \textbf{57.10 ($\pm$1.17)} & \textbf{93.04 ($\pm$0.23)} & \textbf{44.91 ($\pm$2.16)} & \textbf{46.37 ($\pm$1.92)} & \textbf{90.62 ($\pm$1.53)} & \textbf{74.33 ($\pm$0.66)} & \textbf{74.45 ($\pm$0.64)} & \textbf{94.90 ($\pm$0.97)} \\
	            \bottomrule
        \end{tabular}
}
\label{tab:quantitative_results}
\end{table*}

\smallskip
\noindent 
\textbf{Compared Methods.}$~~$
We select the following algorithms to compare with our method. 
\begin{enumerate}
\item \textbf{Kmeans}: An unsupervised clustering algorithm which takes term embedding as features and returns detected synonym sets as clusters. This algorithm requires a predefined cluster number $K$ and we set its value to the oracle number of clusters for each dataset. 

\item \textbf{Louvain \cite{Blondel2008FastUO}}\footnote{\scriptsize \url{https://github.com/taynaud/python-louvain}}: An unsupervised community detection algorithm which takes a graph as input and returns discovered graph communities. To apply this algorithm, we first construct a term graph where each node represents a term. Then, we calculate the cosine similarity between each pair of term embeddings, and if the similarity is larger than a threshold $\alpha$, we will add an edge into the graph. We tune this threshold $\alpha$ on training set. 

\item \textbf{SetExpan
\cite{Shen2017SetExpanCS}\footnote{\scriptsize \url{https://github.com/mickeystroller/SetExpan}}+Louvain}: A two-phase unsupervised approach that first uses SetExpan (\ie, a weakly-supervised set expansion algorithm) to find each term's $k$ nearest neighbors in embedding space, and then construct a $k$-NN graph on which the above Louvain algorithm is applied. We tune the variable $k$ on training set. 

\item \textbf{COP-Kmeans \cite{Wagstaff2001ConstrainedKC}}\footnote{\scriptsize \url{https://github.com/Behrouz-Babaki/COP-Kmeans}}: A semi-supervised variation of the Kmeans algorithm that can incorporate pairwise constraints (\eg, two elements must or cannot be clustered together) and output clusters that satisfy all constraints. We convert training synonym sets into these pairwise constraints and set the oracle number of clusters $K$ for each dataset.

\item \textbf{SVM\footnote{\scriptsize \url{http://scikit-learn.org/stable/modules/svm.html}}+Louvain}: A two-phase supervised approach which first uses a SVM for synonym pair prediction and then groups all predicted pairs into a graph where the Louvain algorithm is applied. The SVM is learned on training set. 

\item \textbf{L2C \cite{Hsu2017LearningTC}\footnote{\scriptsize \url{https://github.com/GT-RIPL/L2C}}}: A supervised clustering method that learns a pairwise similarity prediction neural network and a constrained clustering network on training synonym sets, then applies the learned networks on test vocabulary to detect new entity synonym sets. 

\item \textbf{\SynSetMine}: Our proposed approach which trains a set-instance classifier and integrates it seamlessly into an efficient set generation algorithm.
\end{enumerate}

\smallskip
\noindent 
\textbf{Parameter Settings and Initial Term Embedding.}$~~$
For a fair comparison, we use the same 50-dimension term embedding, trained on each corpus, across all compared methods. 
We tune hyper-parameters in all (semi-)supervised algorithms using 5-fold cross validation on training set. 
For \SynSetMine, we use a neural network with two hidden layers (of sizes 50, 250) as embedding transformer, and another neural network with three hidden layers (of sizes 250, 500, 250) as post transformer (c.f. Figure~\ref{fig:architecture}). 
We optimize our model using Adam with initial learning rate 0.001 and apply dropout technique with dropout rate 0.5. 
For the set generation algorithm, we set the probability threshold $\theta$ be 0.5.
We will discuss the influence of these hyper-parameters later.

\smallskip
\noindent 
\textbf{Evaluation Metrics.}$~~$
As all compared methods output entity synonym sets in the form of clusters, we evaluate them using three standard clustering evaluation metrics.
\begin{itemize}
\item \textbf{ARI} measures the similarity of two cluster assignments. Given the ground truth cluster assignment $\mathcal{C^{*}}$ and model predicted cluster assignment $\mathcal{C}$, we use $TP$ ($TN$) to denote the number of element pairs that are in the same (different) cluster(s) in both $\mathcal{C^{*}}$ and $\mathcal{C}$, respectively. We denote the total number of element pairs in $\mathcal{C^{*}}$ as $N$, and then calculate ARI as follows:
\begin{displaymath}
\small
\text{ARI} = \frac{\text{RI} - \mathbb{E}(\text{RI})}{\max(\text{RI}) - \mathbb{E}(\text{RI})},  \quad \text{RI} = \frac{(TP+TN)}{N},
\end{displaymath}
where $\mathbb{E}(\text{RI})$ is the expected RI of random assignments.

\item \textbf{FMI} is another similarity measure of two cluster assignments. Besides the above $TP$, we use $FP$ ($FN$) to denote the number of element pairs that belong to the same clusters in $\mathcal{C^{*}}$ ($\mathcal{C}$) but in different clusters in $\mathcal{C}$ ($\mathcal{C^{*}}$), respectively. Then, we calculate FMI as follows:
\begin{displaymath}
\small
\text{FMI} = \frac{TP}{\sqrt{(TP+FP)(TP+FN)}}.
\end{displaymath}

\item \textbf{NMI} calculates the normalized mutual information between two cluster assignments. This metric is widely used in literature and its calculation details can be found in \cite{Nguyen2009InformationTM}. 
\end{itemize}

%% Experimental Results
\subsection{Experimental Results}

%% Comparison-1
\smallskip
\noindent 
\textbf{Clustering Performance.}$~~$
We first compare all methods in terms of the clustering performance. 
Results are shown in Table \ref{tab:quantitative_results}.
We can see that overall \SynSetMine outperforms baseline methods across all three datasets. 
The main disadvantage of unsupervised methods (Kmeans, Louvain, and SetExpan+Louvain) is that they cannot utilize supervision signals, which limits their performance when supervision is available.
Compared with Kmeans, COP-Kmeans leverages additional supervision information from the training set and thus has an improvement in performance. 
However, the incorporation of supervision signals is not always straightforward. 
We find that the SVM+Louvain method fails to capture the synonymy relation, probably due to the limited expressive power of SVM. 
Also, the supervised clustering method L2C also does not work very well on synonym datasets regarding both efficiency and the quality of returned clusters.
The reason is that during the training stage, L2C needs to calculate each example's class distribution, and this computation effort is proportional to the training cluster number which is in the scale of thousands\footnote{For comparison, L2C originally runs on MNIST and CIFAR-10 datasets where examples with the same class label is viewed as a cluster and the cluster number is of scale 10-100.}.
Another major deficiency of both SVM+Louvain and L2C is that their learned model is based on pairwise similarity and does not have a holistic view of set. 
On the other hand, \SynSetMine encodes synonym sets directly and therefore is able to capture set-level features beyond pairwise similarity.

%% Comparison-2
\smallskip
\noindent 
\textbf{Set-instance Pair Prediction Performance.}$~~$
To further analyze the importance of encoding set structure, we compare our method with the approach that averages instance-instance pair prediction results, as shown in Figure~\ref{fig:pair-pair}.
Specifically, given a trained set-instance classifier $f(S, t)$, we first construct an instance-instance classifier $g(t_1, t_2)$ to determine whether two instances should be put into the same set. 
We let $g(t_1, t_2)$ to be the mean of $f(\{t_1\}, t_2)$ and $f(\{t_2\}, t_1)$.
Then, for each set-instance pair $(S, t)$, we apply the classifier $g$ to obtain all instance pair prediction results (\ie, $g(t', t), \forall t' \in S$) and average them into the final prediction. 

\begin{figure}
\subfigure[]{
\includegraphics[width = 0.224\textwidth]{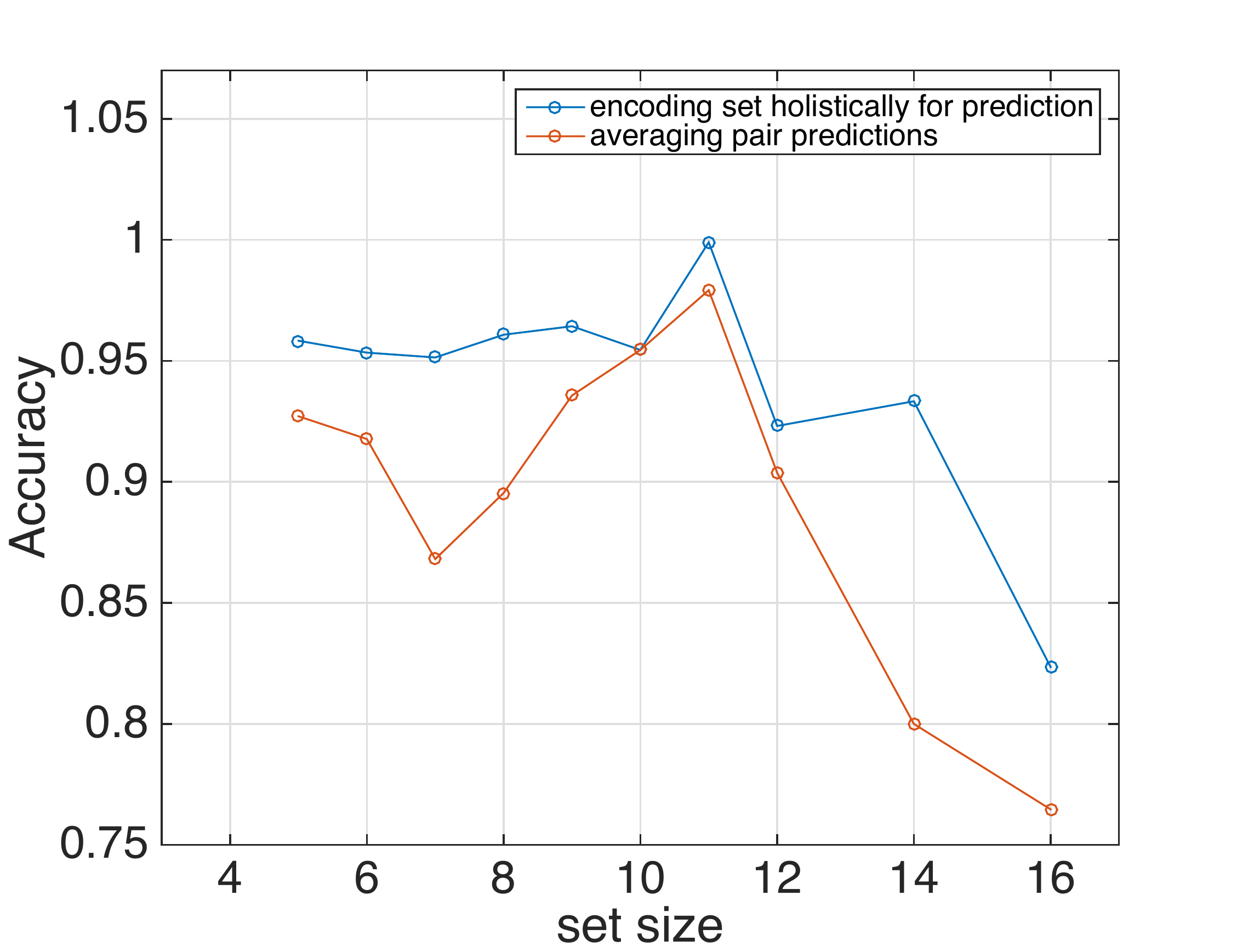}
}
\subfigure[]{
\includegraphics[width = 0.224\textwidth]{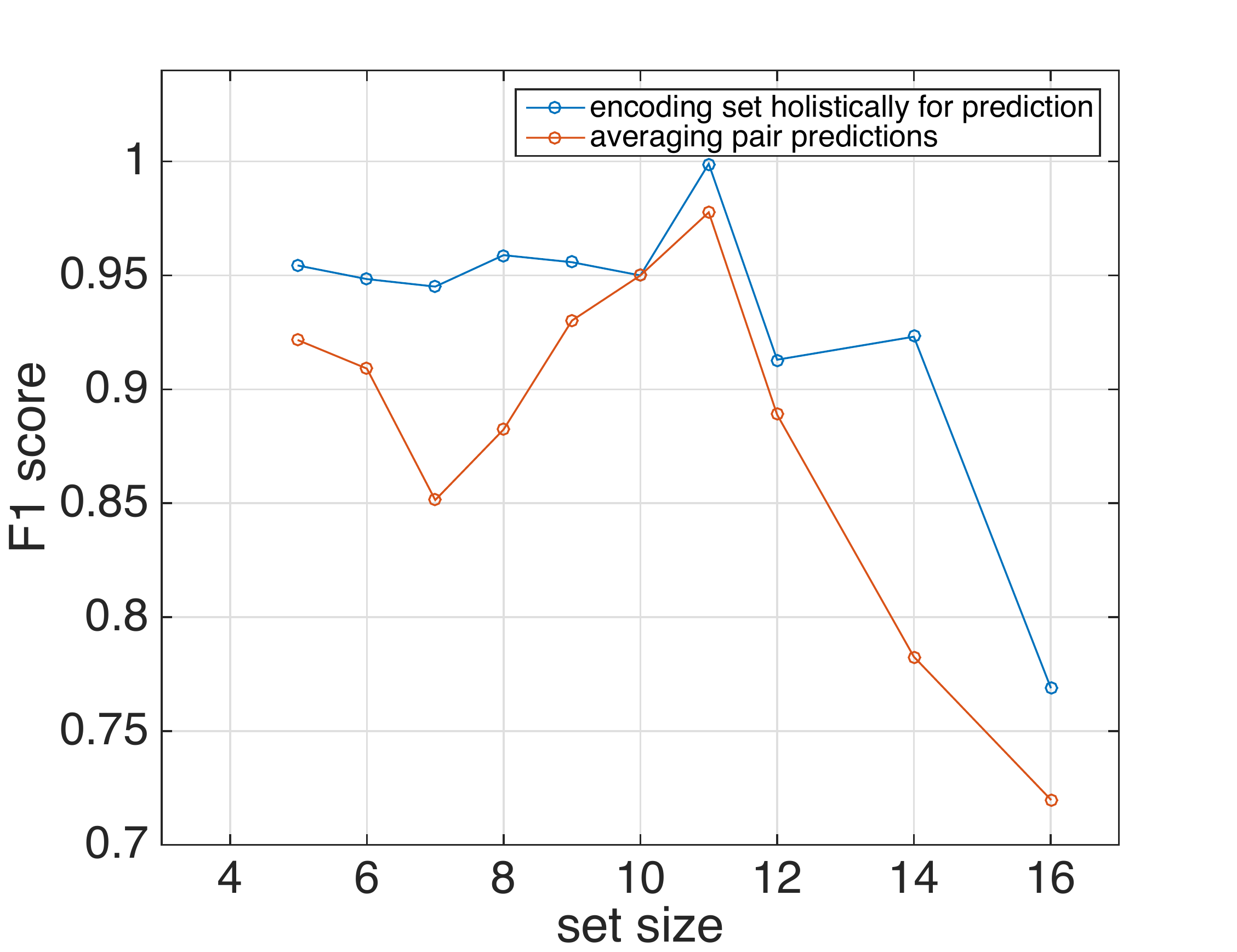}
}
\caption{Evaluation of set-instance classifiers on PubMed dataset using (a) Accuracy and (b) F1 score.}
\label{fig:set_encoding_vs_pairs}
\end{figure}

For evaluation, we convert the testing synonym sets in PubMed dataset into a collection of 3486 set-instance pairs, among which 1743 pairs are positive. 
Results are shown in Figure~\ref{fig:set_encoding_vs_pairs}.
We can see clearly that the set-encoding based approach performs much better than the ``averaging pair prediction'' approach in terms of both accuracy and F1 score. 
This demonstrates the importance of capturing entity relations among the set and modeling the set structure holistically. 

%% Figure: Set encoding effectiveness

%% Comparison-3
\smallskip
\noindent 
\textbf{Efficiency Analysis.}$~~$
We implement our model based on the PyTorch library, same as the L2C baseline. 
We train two neural models (\SynSetMine and L2C) on one Quadro P4000 GPU and run all the other methods on CPU. 
Results are shown in Table \ref{tab:efficiency}. 
Compared with the other supervised neural model L2C, our method can predict significantly faster. 
In addition, our set generation algorithm avoids explicitly computing all pairwise term similarities and thus is more efficient during the prediction stage. 

\begin{table}[!t]
\centering
\caption{Efficiency analysis. Model and data loading time are excluded. GPU time are marked with $^{\star}$. We use ``--'' to indicate that either there is no training time for unsupervised methods, or the model is too slow to train and thus we don't get one trained model for prediction.}
\scalebox{0.67}{
        \begin{tabular}{c|ccc|ccc}
            \toprule
            \multirow{2}{*}{\textbf{Method}} & \multicolumn{3}{c}{\textbf{Training}} & \multicolumn{3}{c}{\textbf{Prediction}} \\
            \cmidrule{2-7}
                           & Wiki & NYT & PubMed         & Wiki & NYT  & PubMed \\
                          \midrule
                        	Kmeans & -- & --  & --  & 1.82s   & 0.88s  & 2.95s \\
                         Louvain & -- & --  & --  & 3.94s  &  20.59s & 74.6s \\
                         SetExpan+Louvain & -- & --  & --  & 323s  & 120s  & 4143s \\
			COP-KMeans & -- & -- & -- & 249s & 37.94s & 713s \\
			\midrule
                         SVM+Louvain & 4.9m & 37s & 1.3h & 29.21s & 5.80s & 101.32s \\
                         L2C & 16.8h$^{\star}$ & 30.7m$^{\star}$  & $>$120h$^{\star}$  & 20.9m$^{\star}$  & 56.6s$^{\star}$  & -- \\
                         \midrule
                         \SynSetMine & 48m$^{\star}$ & 6.5m$^{\star}$ & 7.5h$^{\star}$  & 0.852s$^{\star}$  & 0.348s$^{\star}$  & 1.84s$^{\star}$ \\ 
	            \bottomrule
        \end{tabular}
}
\label{tab:efficiency}
\end{table}

%% Model Analysis and Cases Studies
\subsection{Model Analysis and Cases Studies}

Following we conduct more experiments to analyze each component of our \SynSetMine framework in more details and show some concrete case studies. 

\begin{figure}
\subfigure[Negative sample size]{
\includegraphics[width = 0.21\textwidth]{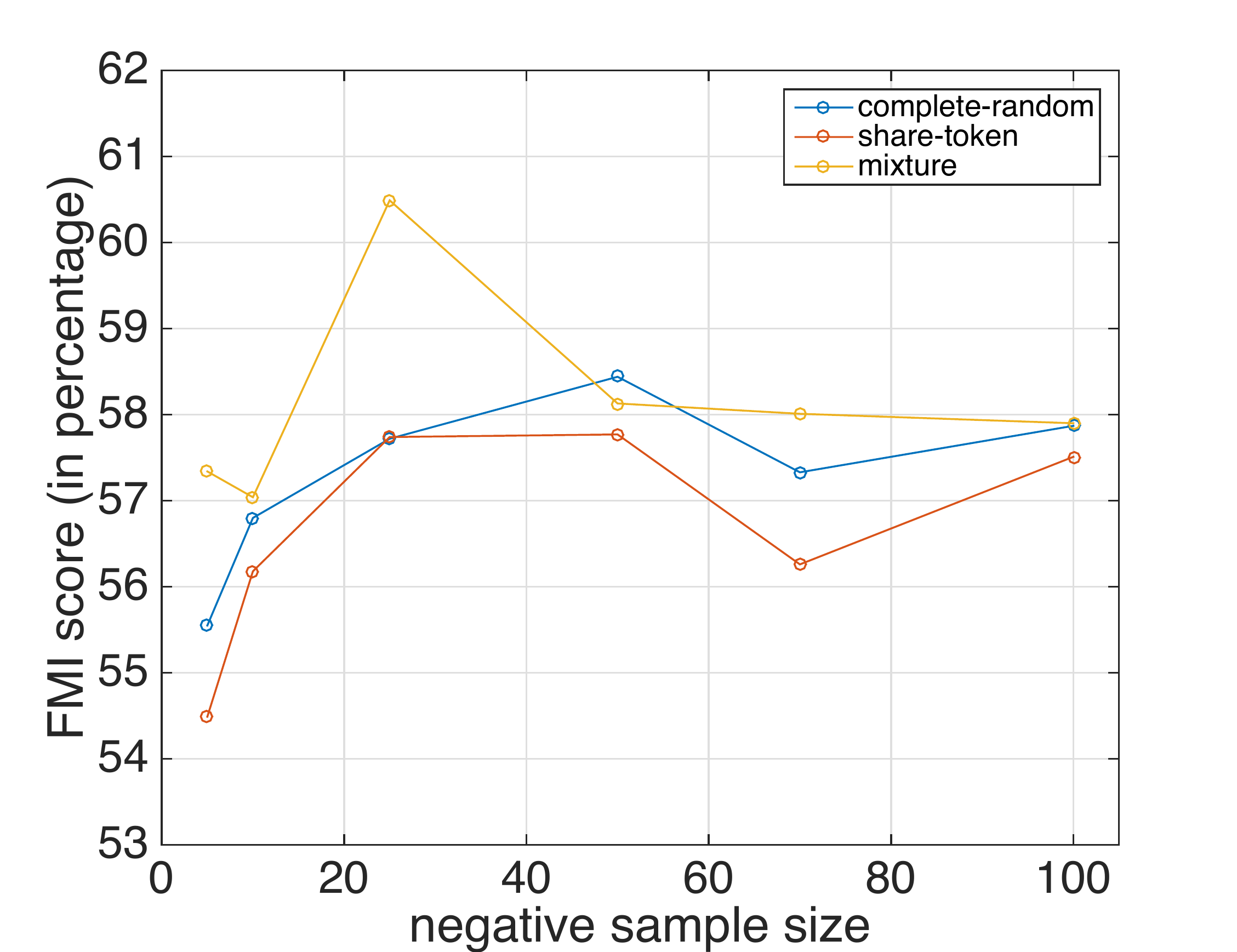}
\label{subfig:negative_sampling}
}
\subfigure[Probability threshold $\theta$]{
\includegraphics[width = 0.21\textwidth]{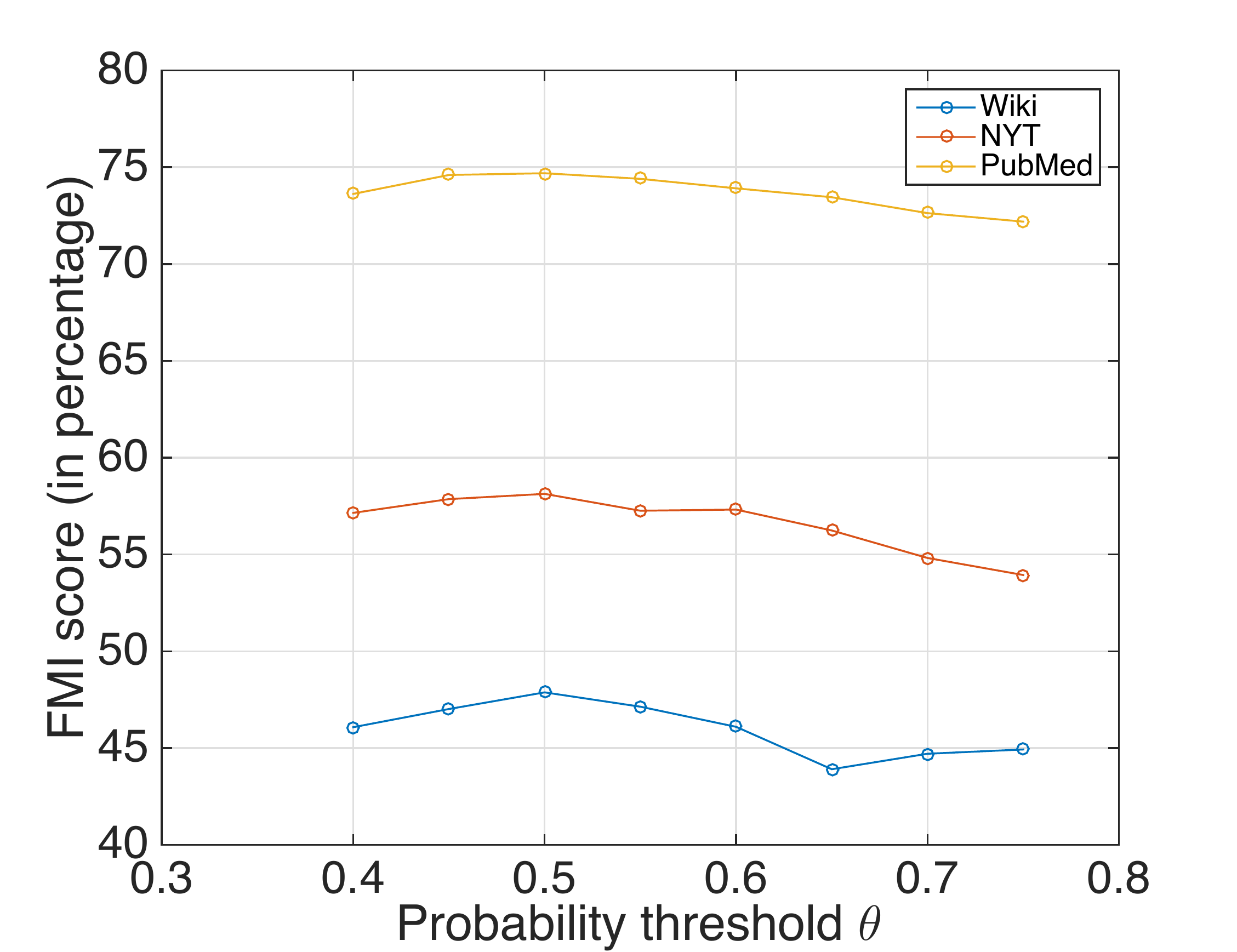}
\label{subfig:threshold_theta}
}
\caption{Hyper-parameters analysis.}
\label{fig:hyper_param_analysis}
\end{figure}

\smallskip
\noindent 
\textbf{Effect of training set-instance pair generation strategy and negative sample size.}$~~$
In order to train the set-instance classifier, we need to first convert the training entity synonym sets (obtained from distant supervision) to a collection of set-instance pairs. 
We study how such conversion strategy and different negative sample sizes will affect the model performance on Wiki dataset. 
Results are shown in Figure~\ref{subfig:negative_sampling}.
First, we find that the \texttt{complete-random} strategy actually performs better than the \texttt{share-token} strategy.
One possible explanation is that the \texttt{complete-random} strategy can generate more \emph{diverse} negative samples and thus provide more supervision signals. 
Second, we observe that by combining the above two strategies and generating more mixed negative samples, our model can be further improved, which again may contribute to the \emph{diversity} of negative samples.

\smallskip
\noindent 
\textbf{Effects of different set-scorer architectures.}$~~$
To demonstrate the necessity of our model components, we first compare the current set-scorer architecture (c.f. Figure \ref{fig:architecture}) with its two ablations. 
Specifically, we leave either the Embedding Transformer (ET) or the Post Transformer (PT) out, and test the performance of remaining models. 
As shown in Table \ref{tab:set_scorer_architecture}, both ET and PT are essential to our model, and removing either of them will significantly damage our model's performance. 
Furthermore, the Post Transformer, which operates on the set representation, is particularly important to our model. 

To further explore the effect of above two transformers, we train our model with different hidden layer sizes. 
Specifically, we use \texttt{Both-X-Y} to denote a set-scorer composed of an Embedding Transformer (of hidden sizes \{50, \texttt{X}\}) and a Post Transformer (of hidden sizes \{\texttt{X, Y, X}\}).
Results are shown in Table \ref{tab:set_scorer_architecture}.
We discover that the performance first keeps increasing when the hidden layer size grows and then drops slightly. 
Also, the best model sizes are consistent for both Wiki and NYT datasets.

%% Table: set scorer architecture
\begin{table}[!t]
\centering
\caption{\small Analysis of set scorer architecture. ``No-ET'' means no Embedding Transformer (ET) module, ``No-PT'' means no Post Transformer (PT) module, and \texttt{Both-X-Y} stands for using both ET (of hidden sizes \{50, X\}) and PT (of hidden sizes \{X, Y, X\}).}
\scalebox{0.75}{
        \begin{tabular}{c|ccc|ccc}
            \toprule
            \multirow{2}{*}{\textbf{Method}} & \multicolumn{3}{c}{\textbf{Wiki}} & \multicolumn{3}{c}{\textbf{NYT}} \\
            \cmidrule{2-7}
                           & ARI & FMI & NMI         & ARI & FMI  & NMI \\
                          \midrule
                        	No-ET & 46.48	& 47.23 & 91.57  & 39.86  & 42.67  & 90.46 \\
                         No-PT & 1.50 & 0.50 & 89.95  & 0.82  & 1.70  & 82.20 \\
			\midrule
                         \texttt{Both-100-200} & 49.38 & 49.56  & 91.21  & 37.64  & 39.37  & 89.33 \\
                         \texttt{Both-150-300} & 53.06 & 53.27  & 91.96  & 43.20  & 44.08  & 89.57 \\
                         \texttt{Both-200-400} & 53.82 & 53.99  & 92.36  & 47.03  & 49.65  & 91.00 \\
                         \texttt{Both-250-500} & \textbf{57.34} & \textbf{58.13}  & \textbf{93.10}  & \textbf{48.89}  & \textbf{51.33}  & \textbf{91.19} \\
                         \texttt{Both-300-600} & 56.26 & 56.51  & 92.92  & 46.65  & 47.30  & 90.01 \\
                         \texttt{Both-350-600} & 55.93 & 56.10  & 92.69  & 47.40  & 48.37  & 90.14 \\
	            \bottomrule
        \end{tabular}
}
\label{tab:set_scorer_architecture}
\end{table}

\smallskip
\noindent 
\textbf{Effect of probability threshold $\theta$.}$~~$
As shown in Algorithm~\ref{alg:algorithm1}, our set generation algorithm requires an input probability threshold $\theta$ to determine whether a term should be added into one of the already detected sets. 
Intuitively, the higher this threshold $\theta$ is, the more conservative our set generation algorithm will be and more sets will be generated. 
In all the above experiments, we set the threshold $\theta$ to be 0.5. 
In this experiment, we intend to empirically study how this hyper-parameter will influence the clustering performance and how \SynSetMine is sensitive to the choice of $\theta$.
Therefore, we run our set generation algorithm with fixed set-instance classifier and varied thresholds. 
The results are shown in Figure \ref{subfig:threshold_theta}. 
First, we notice that the performance of our clustering algorithm is insensitive to $\theta$ and a value within 0.4 and 0.6 is generally good for $\theta$. 
Second, we find that setting $\theta$ to be 0.5 is robust and works well across all three datasets. 

%% Figure: Hyper-parameters analysis

\smallskip
\noindent 
\textbf{Case Studies.}$~~$
Table~\ref{tab:case_studies} presents some example outputs of \SynSetMine.
We can see that our method is able to detect different types of entity synonym sets across different domains. 
Then, in Table~\ref{tab:set_prediction_result}, we show a concrete case comparing our set-instance classifier with the approach that aggregates the instance-instance pair predictions.
Clearly, the set encoding based method can detect more accurate entity synonyms.

%!TEX root = main.tex
% UTF-8 encoding
\section{Related Work}\label{sec:related_work}
There are two lines of related work, including entity synonym discovery and set representation learning. 

\smallskip
\noindent
\textbf{Entity Synonym Discovery. }
Most of previous efforts on entity synonym discovery focus on discovering entity synonyms from (semi-)structured data such as web table schemas \cite{Cafarella2008WebTablesET} and query logs \cite{Chaudhuri2009ExploitingWS,Ren2015SynonymDF}.
In this work, we aim to mine entity synonym sets directly from raw text corpus, which has a border application scope.
Given a corpus, one can leverage co-occurrence statistics \cite{Baroni2004UsingCS}, textual pattern \cite{Nakashole2012PATTYAT,Yahya2014ReNounFE}, distributional similarity \cite{Pantel2009WebScaleDS,Wang2015SolvingVC}, or their combinations \cite{Qu2017AutomaticSD} to extract synonyms.
These methods, however, only find synonymous \emph{term pairs} or \emph{a rank list} of query entity's synonym, instead of entity synonym sets.
Some work attempt to further cut-off the rank list into a set output \cite{Ren2015SynonymDF} or to build a synonym graph and then apply graph clustering techniques to derive synonym sets \cite{Oliveira2014ECOAO,Ustalov2017WatsetAI}.
However, these two-phase approaches suffer from the noise accumulation problem, and cannot directly model the entity set structure.
Comparing to them, our approach can model entity synonym sets holistically and capture important relations among terms in the set.
Finally, there exist some studies, such as finding lexical synonyms from dictionary \cite{Ustalov2017FightingWT} or attribute synonyms from query logs \cite{He2016AutomaticDO}, but this work focuses on mining entity synonyms from raw text corpora. 

%% Table: Case studies: Example Outputs
\begin{table}[!t]
\centering
\caption{Example outputs on three datasets.}
\scalebox{0.7}{
        \begin{tabular}{c|c|c}
            \toprule
	    \textbf{Dataset} &  \textbf{Distant Supervision} &  \textbf{Discovered Synonym Sets} \\
	     \midrule
	     
	    \multirow{2}{*}{\textbf{Wiki}}  & \{\mquote{londres}, \mquote{london}\} &  \{\mquote{gas}, \mquote{gasoline}, \mquote{petrol}\} \\
	    	                \cmidrule{2-3}
	                                  		& \{\mquote{mushroom}, \mquote{toadstool}\} & \{\mquote{roman fort}, \mquote{castra}\} \\
             \midrule
             \multirow{3}{*}{\textbf{NYT}}  & \multirow{2}{*}{ \{\mquote{myanmar}, \mquote{burma}\} } & \{ \mquote{royal dutch shell plc}, \\
             						&               &	    \mquote{royal dutch shell}, \mquote{shell} \} \\
			        \cmidrule{2-3}

							& \{\mquote{honda motor}, \mquote{honda}\} & \{\mquote{chief executive officier}, \mquote{ceo} \} \\
							
             \midrule
             \multirow{2}{*}{\textbf{PubMed}}  & \{\mquote{alzheimers disease},  & \{\mquote{dna microarrays}, \mquote{dna chip},  \\
	                                  		 & \mquote{Alzheimer's dementia}\} & \mquote{gene expression array}, \mquote{dna array}\} \\

	     \bottomrule
        \end{tabular}
}
\label{tab:case_studies}
\end{table}

\smallskip
\noindent
\textbf{Set Representation Learning. }
Our work is also related to set representation learning, which aims to construct permutation invariant representations of sets. 
PointNet \cite{Qi2017PointNetDL} models points in a space as sets and uses multi-layer perceptrons and max pooling function to learn their representation.
DeepSets \cite{Zaheer2017DeepS} establishes the general form of permutation invariant functions and proposes to learn these functions using a deep neural network consisting of multi-layer perceptrons and aggregation functions.
AutoEncSets \cite{Hartford2018DeepMO} further extends DeepSets by modeling the interactions across sets. 
In this work, in addition to just representing given sets, we go beyond one step and aim to \emph{predict} sets (\ie, entity synonym sets) from the vocabulary.  

%% Table: Case studies: Set Prediction
\begin{table}[!t]
\centering
\caption{Comparison of set-instance classifier with the approach that aggregates instance-instance pair predictions.} 
\scalebox{0.72}{
        \begin{tabular}{c|c|c}
            \toprule
	\textbf{Method} & \textbf{Set-instance Classifier} & \textbf{Aggregate Pair Predictions} \\
	\midrule
	\textbf{Synonym set} & \{\mquote{u.k.}, \mquote{britain}\} & \{\mquote{u.k.}, \mquote{britain}\} \\
	\midrule
	\multirow{4}{*}{\textbf{Ranked terms}} & \mquote{uk} & \mquote{uk} \\
								 & \mquote{united kingdom} &\mquote{indie} \\
								 & \mquote{great britain} &  \mquote{united kingdom}\\
								&  \mquote{elizabeth ii} & \mquote{america}  \\
	     \bottomrule
        \end{tabular}
}
\label{tab:set_prediction_result}
\end{table}

%!TEX root = main.tex
% UTF-8 encoding
\section{Conclusions}\label{sec:conclusion}
In this paper, we study how to mine entity synonym sets from raw text corpus.
We propose a framework named \SynSetMine which effectively leverages distant supervision from knowledge bases to learn a set-instance classifier and integrates it in an efficient set generation algorithm to detect new entity synonym sets. 
Extensive experiments on three real-world datasets demonstrate both effectiveness and efficiency of our framework. 
In the future, we plan to further integrate set-instance classifier into the set generation algorithm and learn both of them in an end-to-end fashion. 

%!TEX root = main.tex
% UTF-8 encoding

\section*{Acknowledgements}\label{sec:ack}
This research is sponsored in part by U.S. Army Research Lab. under Cooperative Agreement No. W911NF-09-2-0053 (NSCTA), DARPA under Agreement No. W911NF-17-C-0099, National Science Foundation IIS 16-18481, IIS 17-04532, and IIS-17-41317, DTRA HDTRA11810026, and grant 1U54GM114838 awarded by NIGMS through funds provided by the trans-NIH Big Data to Knowledge (BD2K) initiative (www.bd2k.nih.gov). 
Xiang Ren's research has been supported in part by National Science Foundation SMA 18-29268.
We thank Meng Qu for providing the original synonym discovery datasets and anonymous reviewers for valuable feedback. 

\bibliography{cited}
\bibliographystyle{aaai}
\end{document}